\documentclass[numsec,webpdf,modern,medium,namedate]{oup-authoring-template} 

\graphicspath{{Fig/}}

\usepackage[justification=centering]{caption}

\theoremstyle{thmstyleone}%
\theoremstyle{thmstyletwo}%
\theoremstyle{thmstylethree}%

\usepackage[table,xcdraw]{xcolor}
\usepackage{multirow}
\usepackage{graphicx}
\usepackage{tabularx}

\usepackage{booktabs}

\usepackage{lmodern}

\usepackage{fontawesome5}
\makeatletter
\@ifundefined{faCalendarAlt}{\newcommand{\faCalendarAlt}{}}{}
\makeatother

\usepackage{tikz}
\usetikzlibrary{positioning,arrows}

\makeatletter
\providecommand{\@editor}{}
\makeatother

\begin{document}

\appnotes{Research Article}

\firstpage{1}


\title[Automating Historical Insight Extraction from Large-Scale Newspaper Archives via Neural Topic Modeling]{Automating Historical Insight Extraction from Large-Scale Newspaper Archives via Neural Topic Modeling}

\author[1,$\ast$]{Keerthana Murugaraj\ORCID{0009-0008-5100-055X}}
\author[1]{Salima Lamsiyah\ORCID{0000-0001-8789-5713}}
\author[2]{Marten During\ORCID{0000-0001-7411-771X}}
\author[1]{Martin Theobald\ORCID{0000-0003-4067-7609}}

\authormark{Keerthana Murugaraj et al.}

\address[1]{\orgdiv{Department of Computer Science}, \orgname{University of Luxembourg}, \orgaddress{\street{2, place de l’Université}, \postcode{4365}, \state{Esch-Belval Esch-sur-Alzette}, \country{Luxembourg}}}
\address[2]{\orgdiv{Centre for Contemporary \& Digital History (C$^2$DH)}, \orgname{University of Luxembourg}, \orgaddress{\street{11 Porte des Sciences}, \postcode{4366}, \state{Esch-Belval Esch-sur-Alzette}, \country{Luxembourg}}}

\corresp[$\ast$]{Corresponding author. \href{email:keerthana.murugaraj@uni.lu}{{keerthana.murugaraj@uni.lu}}}




\abstract{Extracting coherent and human-understandable themes from large collections of unstructured historical newspaper archives presents significant challenges due to topic evolution, Optical Character Recognition (OCR) noise, and the sheer volume of text. Traditional topic-modeling methods, such as Latent Dirichlet Allocation (LDA), often fall short in capturing the complexity and dynamic nature of discourse in historical texts. To address these limitations, we employ BERTopic. This neural topic-modeling approach leverages transformer-based embeddings to extract and classify topics, which, despite its growing popularity, still remains underused in historical research. Our study focuses on articles published between 1955 and 2018, specifically examining discourse on \textit{nuclear power} and \textit{nuclear safety}. We analyze various topic distributions across the corpus and trace their temporal evolution to uncover long-term trends and shifts in public discourse. This enables us to more accurately explore patterns in public discourse, including the co-occurrence of themes related to nuclear power and nuclear weapons and their shifts in topic importance over time. Our study demonstrates the scalability and contextual sensitivity of BERTopic as an alternative to traditional approaches, offering richer insights into historical discourses extracted from newspaper archives. These findings contribute to historical, nuclear, and social-science research while reflecting on current limitations and proposing potential directions for future work.}

\maketitle

\section{Introduction}
Newspapers have long been a vital primary source for historians, offering key insights into past events, societal trends, and cultural narratives. They help establish historical timelines, verify facts, and identify trends, providing detailed accounts of people, places, and events. Alongside other primary sources, newspapers continue to shape our understanding of history through rich first-hand documentation \citep{10.1145/544220.544222,10.1093/llc/fqaf025}. Traditional methods of analyzing newspapers, such as manual reading and bibliographic approaches relying on subjective expertise, have notable drawbacks, especially when dealing with large volumes of unstructured textual archives. These methods can be time-consuming, prone to human bias, and inefficient for organizing and analyzing extensive collections of documents. To address these challenges, computational approaches have been increasingly employed in recent studies to interpret factual content and perform text classification tasks \citep{10.1093/llc/fqab099,10.1093/llc/fqaf024}. In contrast to traditional methods, text mining techniques based on Natural Language Processing (NLP)—particularly topic modeling—offer a scalable and automated solution for uncovering latent patterns within large corpora, without the need for manual categorization \citep{6a042974c6ec4a4494c81711b929348b,10.1093/llc/fqaf038,10.1093/llc/fqaf037}. Topic modeling identifies latent topics within vast document collections, thereby providing a more efficient, objective, and scalable solution to the challenges historians face in organizing and analyzing their research \citep{yang-etal-2011-topic,10.1093/llc/fqaf036}

In this paper, we explore the effectiveness of BERTopic \citep{Grootendorst2022BERTopicNT}, a state-of-the-art neural-based topic-modeling technique, in analyzing historical newspaper archives. It leverages transformer-based embeddings \citep{Vaswani2017AttentionIA} to capture nuanced relationships between words and topics, outperforming classical topic-modeling methods such as Latent Dirichlet Allocation (LDA) \citep{10.5555/944919.944937}, probabilistic Latent Semantic Analysis (pLSA), and Non-negative Matrix Factorization (NMF) \citep{Lee1999}. By using pre-trained language models, BERTopic has demonstrated superior performance in terms of generating coherent and human-understandable topics, even in large and complex document collections \citep{medvecki2023multilingual,mutsaddi-etal-2025-bertopic}. The contextualized embedding-based representations enable the model to better understand subtle thematic shifts over time, a feature particularly useful for analyzing historical data where linguistic and contextual variations are significant.

Our work builds upon the foundations laid by our previous research \citep{murugaraj-etal-2025-mining}, which compared classical and neural topic models for historical text analysis. While LDA and NMF have been widely used, they often struggle to handle the complexity of historical newspaper archives, which feature diverse vocabulary, linguistic shifts, and rich historical context. In contrast, more recent neural models like BERTopic address these challenges and offer a more robust and scalable approach to topic modeling. This paper further demonstrates how BERTopic can be applied to large collections of historical newspapers, uncovering hidden patterns and enabling a deeper understanding of historical narratives. The key contributions are summarized as follows.
\sloppy
\begin{itemize}
    \item We develop both static and dynamic topic models: static models capture topic representations without considering time, while dynamic models track thematic evolution, providing insights into historical and societal changes.
    \item We conduct a comprehensive evaluation of BERTopic alongside classical topic models, such as LDA and NMF, including their variants, showcasing the effectiveness of neural-based approaches to capture complex patterns in historical texts.
     \item We demonstrate BERTopic’s effectiveness in extracting meaningful insights from unstructured historical text data, highlighting its scalability with large datasets and its efficiency as an alternative to manual document analysis.
    \item We assess BERTopic’s performance through hyperparameter optimization and compare it to default configurations. We also examine its effectiveness using alternative embedding models, such as Jina, for a more in-depth analysis. 
\end{itemize}

This paper provides historians, researchers, and computational social scientists with an effective tool for analyzing large-scale historical data. By uncovering meaningful themes in extensive newspaper collections, researchers can identify long-term trends, changes in public opinion, and important historical events. The neural-based approach makes it easier to work with unstructured data and solves some of the problems found in traditional methods. This work not only improves topic-modeling techniques for historical analysis but also opens up new possibilities for future research in the humanities and social sciences.


\section{Related Work}\label{rw}

\subsection{\textit{Categories of Topic-Modeling Techniques}}

This section categorizes topic-modeling techniques based on their underlying methods and assumptions, highlighting the unique contributions of each. Broadly, these methods fall into three main categories: \textbf{\textit{algebraic topic models}}, \textbf{\textit{probabilistic topic models}}, and \textbf{\textit{neural topic models}}, each of which is briefly discussed below.

\textit{\textbf{Algebraic topic models}} generally rely on linear algebraic techniques, specifically matrix factorization, to uncover hidden themes of textual data. The Latent Semantic Analysis (LSA)\citep{Deerwester1990IndexingBL} framework was introduced in the 1990s based on the singular value decomposition (SVD) of large document-term matrices. Specifically, LSA decomposes a document-term matrix into term-topic, document-topic, and a diagonal matrix indicating the importance of hidden factors (singular values) to uncover the latent topics. However, because the decomposition can produce negative values, understanding the resulting topics becomes challenging and is less intuitive. In 1999, Lee and Seung introduced NMF\citep{Lee1999} to address this issue by incorporating a non-negative constraint during matrix factorization. This innovation led to the adoption of NMF in topic modeling, as the non-negativity constraint enabled the document-term matrix to be decomposed into two non-negative matrices, yielding topics that are more understandable and coherent. Since its introduction, NMF has gained widespread popularity, leading to the development of several methods categorized as separable NMF \citep{article_Gillis,Kumar2012FastCH,article_Gillis_Nicolas,8666058}. However, these early NMF models are still considered shallow in the sense that they capture a word's meaning regardless of the context in which it appears. Recently, the NMF research community has focused on integrating deep learning techniques into NMF. \citep{10.1109/TPAMI.2016.2554555,10.1016/j.neucom.2022.10.002}

\textit{\textbf{Probabilistic topic models}}, such as probabilistic Latent Semantic Analysis (pLSA) and LDA, form the cornerstone of statistical approaches for identifying latent topics within a collection of documents. pLSA, a probabilistic extension to LSA, integrates probability theory to model the relationships between documents and words through hidden variables. Initially introduced in \citep{10.1145/312624.312649, 10.5555/2073796.2073829}, and later formalized by Hofmann \citep{Hofmann1999ProbabilisticLS}, pLSA estimates topic and word distributions from the document-term matrix using the Expectation Maximization (EM) algorithm \citep{10.1111/j.2517-6161.1977.tb01600.x}. However, pLSA has notable limitations: it suffers from overfitting when handling large datasets, as the number of parameters grows linearly with the number of documents. Additionally, pLSA cannot infer meaningful topics for documents outside the training dataset, as it is heavily tuned to the training corpus. The limitations of pLSA led to the development of more robust algorithms, culminating in the introduction of LDA\citep{10.5555/944919.944937}. It is a widely used probabilistic model for discovering hidden topics from large collections of documents. It assumes that each document is a mixture of different topics and that each topic is characterized by a specific set of words. To determine how topics are assigned to documents and how words are distributed within topics, LDA uses two key parameters: $\alpha$ and $\beta$. The $\alpha$ parameter controls how many topics are likely to appear in a document---higher values mean a document covers many topics, while lower values mean it focuses on just a few. The $\beta$ parameter determines how specific or broad a topic is in terms of its word distribution. By using these parameters, LDA helps to identify meaningful themes across documents while ensuring that the model can generalize well to new, unseen data. Many researchers have extended and improved LDA to make topic modeling even more effective \citep{NIPS2005_9e82757e,10.1145/1143844.1143859,NIPS2009_0d0871f0,10.5555/1699510.1699543,Jelodar2017LatentDA,10.1145/3462478}.

\textit{\textbf{Neural topic models}} have been developed to overcome some of the limitations of traditional models like LDA and pLSA. These classical models require predefined topic numbers and rely on Bag-of-Words (BoW) representations, which often fail to capture deeper semantic relationships in text. Additionally, their inference methods, such as Variational Inference \citep{10.5555/944919.944937} and Gibbs Sampling \citep{Gibbs}, are computationally expensive and inefficient for large datasets. To overcome the limitations of traditional topic models, researchers turned to neural networks, which offer several key advantages. First, neural topic models can learn richer representations of text by leveraging word embeddings, capturing semantic relationships that Bag-of-Words (BoW) models like LDA and pLSA often miss. Second, they can scale efficiently to large datasets because deep learning models, especially those based on Variational Autoencoders (VAEs) \citep{Kingma2013AutoEncodingVB} or Transformers \citep{Vaswani2017AttentionIA}, utilize parallel processing on GPUs. This makes training faster compared to traditional methods like Gibbs Sampling or Variational Inference, which are computationally expensive. Lastly, neural models are more flexible, allowing researchers to integrate external knowledge, multimodal data, or hierarchical structures into topic modeling, making them applicable to a wider range of real-world tasks.

A significant milestone in neural topic modeling was achieved by Mikolov et al. in 2013 \citep{NIPS2013_9aa42b31}, who explored the integration of neural networks with topic models. Building on this foundation, \citep{Cao2015ANN} introduced the Neural Topic Model in 2015, offering a more flexible and expressive framework by combining deep learning with the traditional probabilistic models. In 2016, \citep{pmlr-v48-miao16} proposed a neural topic model based on the VAE framework \citep{Kingma2013AutoEncodingVB}, which has since become a widely adopted approach in the field. Further advancements included lda2vec, introduced by \citep{Moody2016MixingDT}, which effectively integrated LDA with word embeddings from the skip-gram model to generate semantically enriched topic representations. In 2017, the Product of Experts Latent Dirichlet Allocation (ProdLDA) model \citep{Srivastava2017AutoencodingVI} incorporated neural networks into the LDA framework by leveraging a VAE for more efficient inference. These developments have paved the way for continuous advancements in neural topic modeling, leading to the exploration of novel architectures and methodologies aimed at improving scalability, coherence, and adaptability across diverse textual corpora.

In recent years, topic-modeling techniques have evolved to incorporate word and document embeddings, improving upon traditional probabilistic models. The Embedded Topic Model (ETM) \citep{dieng-etal-2020-topic}, introduced in 2020, modifies the standard LDA approach by integrating word embeddings directly into the topic-word distribution, allowing for more semantically meaningful topic representations. Around the same time, Top2Vec \citep{Angelov2020Top2VecDR} emerged as another alternative, leveraging document and word embeddings to automatically identify topics without requiring predefined topic numbers. This approach combines the Distributed Bag of Words (DBOW) variant of Doc2Vec with dimensionality reduction via Uniform Manifold Approximation and Projection (UMAP) \citep{McInnes2018UMAPUM} and clustering using Hierarchical Density-Based Spatial Clustering of Applications with Noise (HDBSCAN) \citep{McInnes2017hdbscanHD}. By defining each topic as the centroid of a cluster of document embeddings, Top2Vec identifies key topic words based on their proximity to this central point. However, this method assumes that topics form well-defined clusters with clear centroids, which may not always align with the structure of real-world data, where topics can be irregularly distributed. To overcome these limitations, BERTopic \citep{Grootendorst2022BERTopicNT} was introduced in 2022, incorporating contextualized embeddings and a more refined topic representation strategy. Unlike Top2Vec, which primarily relies on static word embeddings, BERTopic utilizes transformer-based sentence embeddings, such as BERT \citep{devlin-etal-2019-bert}, to capture deeper semantic relationships. Both methods share key steps, including document embedding, dimensionality reduction, and density-based clustering, but BERTopic distinguishes itself through its use of a class-based TF-IDF (C-TF-IDF) technique to extract representative topic words. This enhancement allows BERTopic to generate more coherent and human-understandable topics, making it particularly effective for complex and large-scale text data.


\begin{figure*}[ht]
  \centering
  \begin{tikzpicture}[
      node distance=0.4cm and 0.4cm,
      every node/.style={
        rectangle,
        draw,
        align=center,
        font=\sffamily,
        minimum width=1.9cm,
        minimum height=0.8cm
      },
      core/.style={fill=green!10, draw=green!40!black},
      static/.style={fill=blue!10,  draw=blue!40!black},
      dynamic/.style={fill=orange!10, draw=orange!40!black},
      arrow/.style={->, thick, black}
    ]
    \node[core]   (data)    { \faDatabase \\ Data Collection \\  Gather raw texts};
    \node[core]   (prep)    [below=of data]  { \faCogs \\ Data Preparation\\  Clean \& standardize};
    \node[core]   (embed)   [below=of prep]  { \faCube \\ Precompute Embeddings\\  Encode docs to vectors};
    \node[core]   (latent)  [below=of embed] { \faLightbulb \\ Latent Theme Identification\\ (BERTopic) \\  Discover topic space};

    \node[static] (filterS) [below left=of latent] {\faFilter Static Features \\   Embeddings + Docs + Doc IDs};
    \node[static] (static)  [below=of filterS]     {\faBars \\ Static Topic Modeling\\Extract global Topics};
    \node[static] (eval)    [below=of static]      {\faCheckCircle \ Evaluation\\Assess coherence \\ and diversity};

    \node[dynamic] (filterD) [below right=of latent] { \faFilter Dynamic Features \\   Embeddings + Docs + Doc IDs \\ + year metadata \faCalendarAlt};
    \node[dynamic] (dynamic)[below=of filterD]        { \faChartArea \\ Dynamic Topic Modeling\\  Extract evolving topics};
    \node[dynamic] (viz)    [below=of dynamic]        {\faChartLine  \ Visualization\\Trend over time};

    \draw[arrow] (data)   -- (prep);
    \draw[arrow] (prep)   -- (embed);
    \draw[arrow] (embed)  -- (latent);

    \draw[arrow] (latent) -- (filterS);
    \draw[arrow] (filterS) -- (static);
    \draw[arrow] (static)  -- (eval);

    \draw[arrow] (latent) -- (filterD);
    \draw[arrow] (filterD) -- (dynamic);
    \draw[arrow] (dynamic) -- (viz);
  \end{tikzpicture}
  \caption{Methodology: Overview of the processing pipeline for static and dynamic topic modeling, illustrating each stage from data collection through embedding computation and latent theme identification, with parallel static and dynamic modeling paths.}
  \label{fig:Methodology}
\end{figure*}
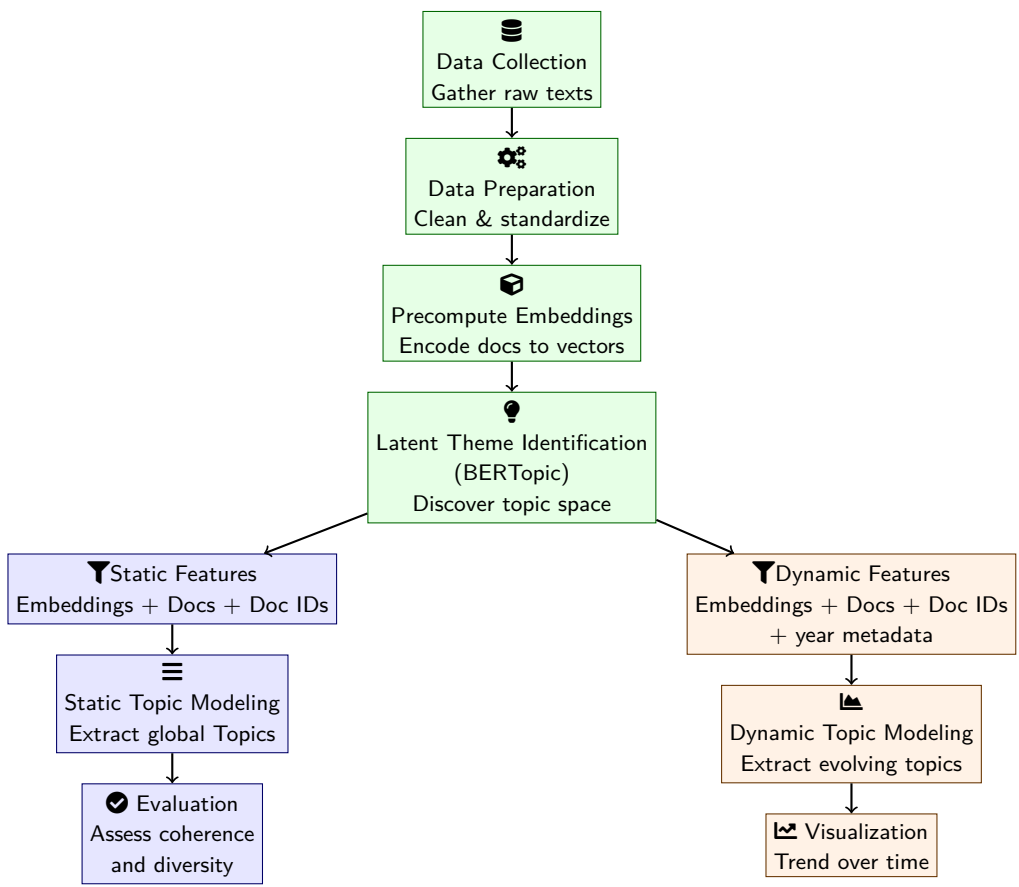


\subsection{\textit{Historical Topic Modeling with Neural Methods}}
In our previous work \citep{murugaraj-etal-2025-mining}, we provided an overview of historical topic-modeling approaches, highlighting the limited use of neural-based models in this domain\citep{ArsenievKoehler2020IntegratingTM,Cvejoski2023NeuralDF,martinelli-etal-2024-exploring,ginn2024historia}, despite their increasing success in various NLP tasks. Historical datasets present unique challenges, such as linguistic variation, thematic evolution, and cultural nuances, that traditional models often fail to address. This gap presents an opportunity to improve topic modeling in this domain using modern neural approaches. Building on our earlier findings, where BERTopic outperformed traditional models like LDA and NMF, this paper further investigates its practical application to real-world historical corpora. We extend our previous work by comparing BERTopic with enhanced variants of LDA and NMF that incorporate features such as TF-IDF weighting and Named Entity Recognition (NER). As large, complex historical archives become increasingly available, there is a growing need for scalable, automated tools that can extract meaningful patterns while preserving historical context. BERTopic offers a compelling solution to this challenge. To the best of our knowledge, this is the first study to apply neural topic modeling to such an extensive collection of historical newspaper archives, setting the stage for future research in historical text analysis using advanced neural methods.  

\section{Methodology}

In this section, we present an overview of the key methodological steps (see Fig.~\ref{fig:Methodology}.) undertaken in our study. These include the collection and curation of a historical newspaper dataset, preprocessing and preparation of the textual data, transformation of documents into dense vector representations using embedding techniques, identification of latent thematic structures through topic modeling, and the optimization of BERTopic’s hyperparameters to enhance the coherence and relevance of extracted topics. Each step is designed to systematically support our goal of uncovering meaningful patterns in large-scale historical text archives.

\subsection{\textit{Data Collection}}
The dataset investigated in this paper is sourced from "impresso: Media Monitoring of the Past"\footnote{\url{https://impresso-project.ch/app/}}, an interdisciplinary project \citep{Ehrmann2020HistoricalNC} that brings together three prominent institutions across Luxembourg and Switzerland, each contributing expertise from distinct but complementary fields: the Digital Humanities Laboratory (DHLAB) at the École Polytechnique Fédérale de Lausanne (EPFL), the Institute of Computational Linguistics (ICL) at the University of Zurich, and the Centre for Contemporary and Digital History (C2DH) at the University of Luxembourg. For this study, we obtained a curated subset of the impresso dataset directly from C2DH. This collection comprises over 148,000 digitized articles and advertisements sourced from historical newspapers published between 1955 and 2018. The subset specifically targets content thematically centered on "\textit{nuclear power}" and "\textit{nuclear safety}", thereby offering a rich corpus for exploring public discourse surrounding nuclear technologies and their societal implications across several decades. Each article in the dataset is associated with a unique identifier (UID), which facilitates precise referencing, tracking, and alignment throughout our preprocessing, embedding, and topic modeling workflows.

\subsection{\textit{Data Preparation and Preprocessing}}
\begin{figure*}[ht!]
\centering
\includegraphics[scale=0.52]{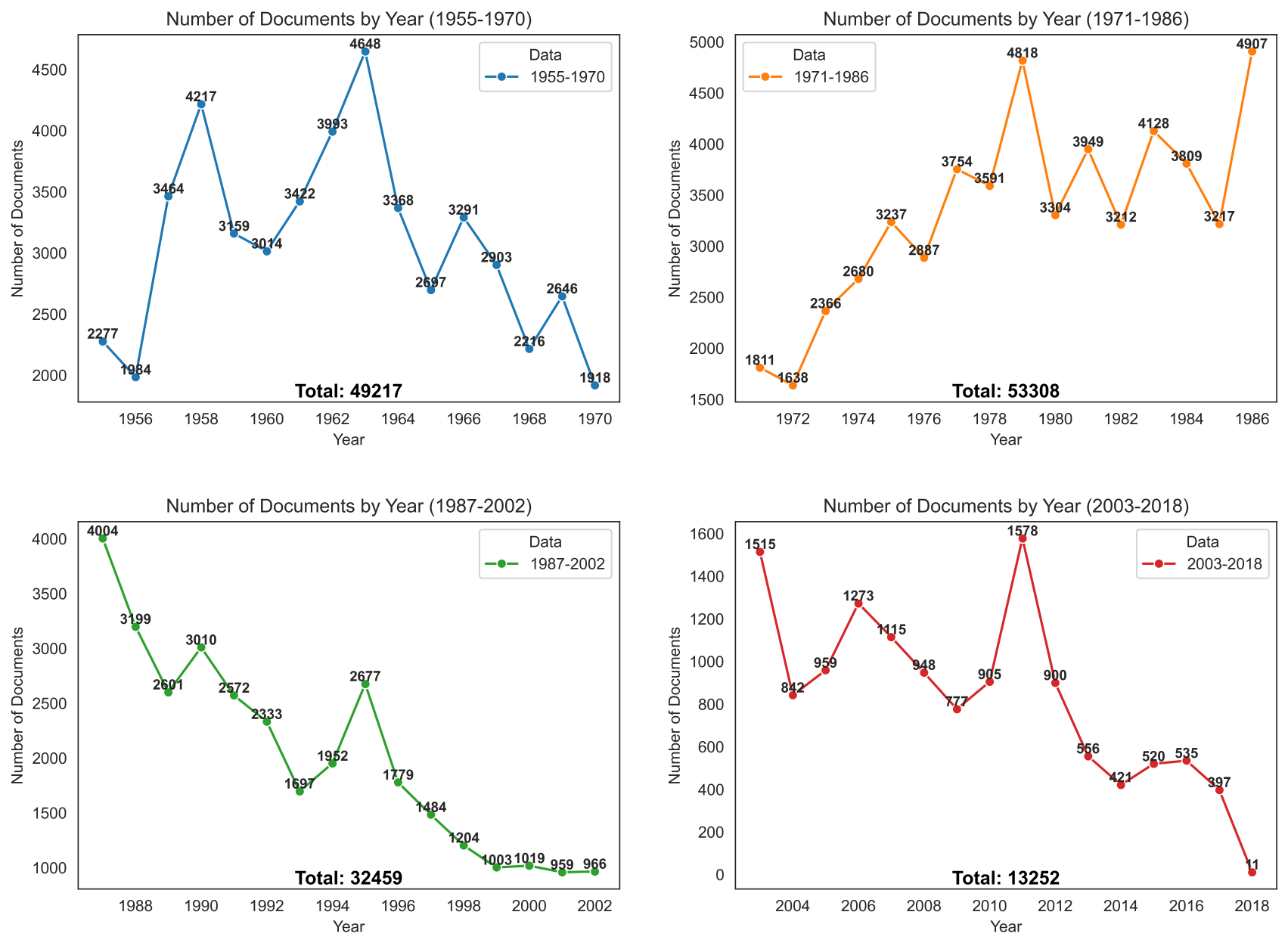}   
\caption{Year-wise distribution of documents across different data subsets, illustrating the temporal spread and volume of content used in our analysis. This visualization helps contextualize topic evolution by showing how data density varies over time.}
\label{fig:doc.dist}
\end{figure*}

The dataset is divided into four subsets based on publication years: 1955–1970, 1971–1986, 1987–2002, and 2003–2018. We conducted a Preliminary Data Exploration (PDE) to gain an initial understanding of the dataset’s characteristics. PDE is a crucial first step, as it helps uncover underlying patterns, detect outliers, identify missing values, and assess the overall structure of the data. As part of the PDE, we examined the distribution of documents across the subsets (see Fig.~\ref{fig:doc.dist}.), analyzed word-frequency patterns, evaluated token counts (minimum and maximum), and checked for missing values. This early analysis is essential to ensure that the data is properly structured, identify potential issues, such as imbalanced document distributions or missing entries, and guide subsequent preprocessing and topic-modeling tasks. PDE allowed us to identify areas of the dataset that required special attention and helped optimize the overall modeling workflow.

Furthermore, the collected dataset comprises French and German newspaper texts, along with some content in Luxembourgish. To ensure consistency in evaluation, all texts were automatically translated into English using Google Translate, a multilingual neural machine-translation service. We manually reviewed the outputs to correct significant translation errors. After translation, the text was cleaned by removing extraneous noise and unwanted characters, including special symbols (e.g. \&, \$, @), using regular expressions.

This step standardized the dataset by eliminating non-textual elements that could interfere with text processing. Next, we removed documents with fewer than 20 tokens and applied linguistic preprocessing using spaCy, a robust NLP library. This included stopword removal to filter out non-informative words, lowercasing for uniformity, and tokenization to segment text into individual tokens. We retained only full stops, removing other punctuation marks to preserve sentence boundaries without introducing noise. Finally, we applied lemmatization to reduce inflected words to their base forms. This preprocessing pipeline ensured a clean, consistent, and well-structured dataset, providing a solid foundation for downstream tasks such as topic modeling.

\subsection{\textit{Document Embeddings}}  
We use Sentence Transformer \citep{Reimers2019SentenceBERTSE} based embeddings to represent documents as dense vectors in a semantic vector space, enabling precise analysis of semantic similarity.
Specifically, we use the \textit{'gte-base-en-v1.5} \& \textit{jina-embeddings-v2-base-en'} models from the Huggingface library, which we selected for their strong performance in capturing nuanced semantic relationships and their ability to process sequences of up to 8,192 tokens. This extended sequence length allows for richer contextual representations, making them well-suited for handling complex and lengthy textual data. Each document is transformed into a dense numerical vector, where semantically similar documents are mapped closer together in the embedding space. These embeddings form the basis for topic modeling, facilitating the identification of coherent clusters of related documents. By leveraging high-quality embeddings, we aim to improve the conceptual clarity and coherence of the extracted topics, ensuring that the derived keywords accurately capture the underlying themes.

\subsection{\textit{Latent Themes Identification}}

We use BERTopic \citep{Grootendorst2022BERTopicNT}, an unsupervised topic-modeling method that combines transformer-based embeddings with hierarchical clustering. Our analysis applies both \textbf{\textit{static}} and \textbf{\textit{dynamic}} BERTopic models to capture latent themes in historical archives. In the \textbf{static topic-modeling approach}, each document is first converted into a numerical representation using precomputed embeddings that preserve semantic meaning. To reduce the high dimensionality while retaining local structure, we further apply UMAP \citep{McInnes2018UMAPUM}, projecting the embeddings into a lower-dimensional space. These reduced embeddings are then clustered using HDBSCAN \citep{McInnes2017hdbscanHD}, which groups similar documents without requiring a predefined number of clusters. Topics are represented using C-TF-IDF (Class-based Term Frequency–Inverse Document Frequency), which highlights the most representative words for each cluster, improving clarity and facilitating more intuitive understanding. In the \textbf{dynamic topic-modeling approach}, we extend BERTopic to track how topics evolve over time. We segment the documents using temporal metadata to observe changes in topic distribution and importance across different periods. To further refine topic representations, we incorporate KeyBERT\citep{grootendorst2020keybert}, which extracts keywords that best capture each topic’s core meaning. By combining static and dynamic modeling, our approach offers a comprehensive view of thematic structures in historical archives, revealing both long-term trends and temporal shifts in topic distributions.

\subsection{\textit{BERTopic Hyperparameter Optimization}} 
\sloppy
To assess whether hyperparameter tuning improves BERTopic’s performance and topic quality, we used the distributed asynchronous Hyperparameter Optimization (HPO) \citep{10.5555/3042817.3042832} framework to fine-tune key hyperparameters from UMAP and HDBSCAN, including \texttt{min\_topic\_size}, \texttt{n\_neighbors}, \texttt{n\_components}, \texttt{min\_cluster\_size}, and \texttt{min\_samples}. More specifically, we employed \textbf{Hyperopt} \citep{Bergstra2013HyperoptAP}, a distributed optimization library, and selected the \textbf{Tree of Parzen Estimators (TPE)} algorithm \citep{NIPS2011_86e8f7ab} for its effectiveness in high-dimensional search spaces. The TPE algorithm works iteratively, starting with random sampling and modeling two distributions, one for high-performing configurations and one for all observations. It then focuses the search on the most promising regions of the parameter space and continues until the predefined number of evaluations is completed.

The Hyperopt framework involves three main steps. First, we define the \textbf{objective function}, which aims to optimize hyperparameters by maximizing \textit{topic coherence} and \textit{topic diversity} to improve the quality of the topic model. Second, we specify the \textbf{search space}, which is the user-defined range of possible values for each hyperparameter that Hyperopt explores to find the best configuration. Third, we select the \textbf{search algorithm} by using the \texttt{fmin} function provided by Hyperopt. This function requires several inputs: the \textbf{objective function} to be minimized or maximized, the \textbf{parameter space} that defines the search domain, and the \textbf{search algorithm}, for which we use \texttt{tpe.suggest}, allowing Hyperopt to choose new hyperparameters based on prior evaluations. Additionally, we set the \textbf{maximum number of evaluations} using the \texttt{max\_evals} parameter and track results with the \textbf{trials object} (\texttt{trials}), which stores the history of all iterations for further analysis. The \texttt{fmin} function returns the best-performing hyperparameter configuration, while the \texttt{trials} object provides detailed insights into the performance of all tested configurations.

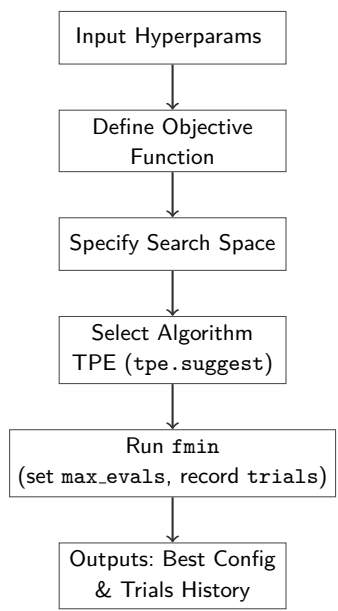
\begin{figure}[ht]
  \centering
  \begin{tikzpicture}[
      node distance=0.6cm,
      every node/.style={
        rectangle,
        draw=white!30!black,
        fill=white!10,
        align=center,
        font=\sffamily,
        minimum width=3cm,
        minimum height=0.7cm
      },
      arrow/.style={->, thick, gray!40!black}
    ]
       \node (input) {
      Input Hyperparams };
    \node (obj)    [below=of input] {
      Define Objective\\      Function
    };
    \node (space)  [below=of obj]   {Specify Search Space};
    \node (alg)    [below=of space] {
      Select Algorithm\\
      TPE (\texttt{tpe.suggest})
    };
    \node (run)    [below=of alg]   {
      Run \texttt{fmin}\\
      (set \texttt{max\_evals}, record \texttt{trials})
    };
    \node (output) [below=of run]   {
      Outputs: Best Config\\
      \& Trials History
    };

    \draw[arrow] (input)  -- (obj);
    \draw[arrow] (obj)    -- (space);
    \draw[arrow] (space)  -- (alg);
    \draw[arrow] (alg)    -- (run);
    \draw[arrow] (run)    -- (output);
  \end{tikzpicture}
  \caption{ Workflow for BERTopic hyperparameter tuning.}
  \label{fig:hyperopt_workflow}
\end{figure}

\section{Experimental Results}
\sloppy
In this section, we present the hardware and software specifications used in our experiments, the methods employed for evaluation, and the results of both static and dynamic topic models. We then provide a detailed discussion and comparison of the performance of these models, highlighting their strengths and limitations.  
\subsection{\textit{Experimental Settings}}
All experiments were conducted on \href{https://luxembourg.public.lu/en/invest/innovation/meluxina-supercomputer.html}{{Meluxina}}, Luxembourg’s cutting-edge supercomputer, one of the most powerful in Europe. It provides robust facilities for high-performance computing (HPC) and AI workloads. For our experiments, we utilized one NVIDIA A100 GPU, the most powerful GPU available for deep-learning tasks on Meluxina. The topic-modeling techniques employed in our studies include Gensim LDA\citep{10.5555/944919.944937}, Gensim Online NMF\citep{7676413}, and BERTopic version 0.16.3. For the default \texttt{BERTopic} parameter settings, we used UMAP for dimensionality reduction with the following configuration: \texttt{n\_neighbors=15}, \texttt{n\_components=5}, \texttt{min\_dist=0.0}, and \texttt{metric="cosine"}. Clustering was performed using HDBSCAN with \texttt{min\_cluster\_size=15}, \texttt{metric="euclidean"}, \texttt{cluster\_selection\_method="eom"}, and \texttt{prediction\_data=True}. In the hypertuned \texttt{BERTopic} configuration, the best-performing hyperparameters were: \texttt{min\_cluster\_size=31}, \texttt{min\_samples=24}, \texttt{min\_topic\_size=37}, \texttt{n\_components=5}, and \texttt{n\_neighbors=10}.

\subsection{\textit{Baseline Methods}}
In our previous work \citep{murugaraj-etal-2025-mining}, we evaluated BERTopic using various embedding models and selected the best-performing one for our topic-modeling tasks. After identifying the optimal embedding model, we further compared BERTopic with the recent Jina embedding model. In addition, we evaluated BERTopic against classical models, namely LDA and NMF, along with their respective variants, across all data subsets. For LDA, we began with the standard implementation using a Bag-of-Words (BoW) corpus representation. We then created two variants: one using a TF-IDF corpus and another incorporating named entity recognition (NER). This resulted in three LDA models for comparison: \textit{LDA}, \textit{LDA-tfidf}, and \textit{LDA-ner}. Traditional LDA identifies latent topics by analyzing word distributions across documents. In contrast, LDA-ner incorporates named entities (e.g., person names, locations, organizations) as input features, enhancing the model's ability to identify real-world concepts that are crucial for topic understanding. We applied the same methodology to NMF. Starting with the standard model, we developed two additional variants: NMF with TF-IDF input and NMF with NER-extracted entities. This produced three models: \textit{NMF}, \textit{NMF-tfidf}, and \textit{NMF-ner}.


\definecolor{lightyellow}{rgb}{1.0, 1.0, 0.80} 
\definecolor{lightblue}{rgb}{0.80, 0.90, 1.0}   
\definecolor{lightgreen}{rgb}{0.80, 1.0, 0.80}  
 
\begin{table*}[ht!]
    \centering
    \small
    \renewcommand{\arraystretch}{0.9}
    \setlength{\tabcolsep}{3.5pt}
    \begin{tabular}{lccccccccccccc}
        \toprule
        \textbf{Model} & \textbf{\#T} & \multicolumn{3}{c}{\textbf{1955--1970}} & \multicolumn{3}{c}{\textbf{1971--1986}} & \multicolumn{3}{c}{\textbf{1987--2002}} & \multicolumn{3}{c}{\textbf{2003--2018}} \\
        & & \textit{TC} & \textit{TD} & \textit{Time} & \textit{TC} & \textit{TD} & \textit{Time} & \textit{TC} & \textit{TD} & \textit{Time} & \textit{TC} & \textit{TD} & \textit{Time} \\
        \midrule
        \multicolumn{14}{c}{\textbf{Classical Models}} \\
        \midrule
        \multirow{5}{*}{LDA} & 10 & 0.10 & 0.78 & 27.79 & 0.09 & 0.74  & 26.68 & 0.09 & 0.74 & 17.87 & 0.04 & 0.68 & 6.68 \\
        & 20 & 0.07 & 0.74 & 53.10 & 0.05 & 0.74 & 50.70 & 0.08 & 0.76 & 26.70 & 0.08 & 0.68 & 10.42 \\
        & 30 & 0.09 & 0.78 & 82.34 & 0.04 & 0.76 & 51.09 & 0.05 & 0.77 & 32.88 & 0.05 & 0.70 & 12.81 \\
        & 40 & 0.03 & 0.75 & 106.05 & 0.05 & 0.75 & 65.91 & 0.05 & 0.74 & 42.62 & -0.01 & 0.65 & 16.15 \\
        & 50 & 0.03 & 0.80 & 87.89 & 0.13 & 0.79 & 63.79 & 0.05 & 0.74 & 42.62 & -0.01 & 0.66 & 16.28 \\
        \midrule
        \multirow{5}{*}{LDA-tfidf} & 10 & -0.16 & 0.99 & 24.81 & -0.16 & 1.0 & 21.73 & -0.19 & 1.0 & 15.00 & -0.15 & 0.92 & 6.61  \\
        & 20 & -0.16 & 0.93 & 38.90 & -0.20 & 0.99 & 32.35 & -0.16 & 0..98 & 22.24 & -0.17 & 0.92 & 8..86 \\
        & 30 & -0.16 & 0.94 & 44.35 & -0.18 & 0.97 & 30.22 & -0.19 & 0.97 & 26.19 & -0.18 & 0.91 & 10.51\\
        & 40 & -0.18 & 0.95 & 51.05 &  -0.18 & 0.97 & 45.50 & -0.18 & 0.98 & 31.09 & -0.15 & 0.93 & 15.40 \\
        & 50 & -0.18 & 0.95 & 50.01 & -0.18 & 0.94 & 46.39 & -0.16 & 0.97 & 30.43 & -0.14 & 0.94 & 12.07\\
        \midrule
         \multirow{5}{*}{LDA-ner} & 10 &  0.07 & 0.8 & 1190.71 & 0.07 & 0.69 & 975.53 & 0.08 & 0.76 & 639.93 &  0.08 & 0.71 & 1188.93 \\
        & 20 &  0.06 & 0.77 & 1198.90 & 0.09 & 0.73 & 981.66 & 0.05 & 0.72 & 643.42 & 0.06 & 0.67 & 268.06\\
        & 30 & 0.08 & 0.80 & 1214.29 & 0.09 & 0.73 & 991.84 & 0.06 & 0.73 & 652.500 & 0.05 & 0.67 & 244.65  \\
        & 40 & 0.03 & 0.77 & 1224.99 & 0.06 & 0.76 & 2796.78 & 0.03 & 0.72 & 659.85 & 0.01 & 0.64 & 246.97  \\
        & 50 & 0.03 & 0.75 & 1218.02 & 0.05 & 0.79 & 5931.39 & 0.03 & 0.76 & 669.50 & -0.01 & 0.62 & 319.80 \\
        \midrule
        \multirow{5}{*}{NMF} & 10 & 0.08 & 0.77 & 93.80 & 0.08 & 0.71 & 92.59 & 0.08 & 0.81 & 49.59 & 0.08 & 0.76 & 20.78 \\
        & 20 & 0.08 & 0.65 & 198.03 & 0.09 & 0.68 & 202.68 & 0.10 & 0.73 & 998.24 & 0.08 & 0.60 & 37.28 \\
        & 30 & 0.08 & 0.56 & 229.83 & 0.09 & 0.60 & 286.73 & 0.11 & 0.65 & 114.92 & 0.10 & 0.52 & 53.83 \\
        & 40 & 0.09 & 0.53 & 599.10 & 0.09 & 0.55 & 380.56 & 0.12 & 0.62 & 157.97 & 0.11 & 0.54 & 70.07 \\
        & 50 & 0.08 & 0.49 & 685.23 & 0.09 & 0.54 & 486.18 & 0.10 & 0.55 & 234.19 & 0.12 & 0.47 & 113.99 \\
        \midrule
           \multirow{5}{*}{NMF-tfidf} & 10 & 0.11 & 0.93 & 270.62 & 0.13 & 0.91 & 109.84 & 0.08 & 0.71 & 649.24 &  0.16 & 0.93 & 22.99  \\
        & 20 & 0.12 & 0.84 & 1034.23 & 0.12 & 0.83 & 224.18 & 0.11 & 0.72 & 698.68 & 0.19 & 0.78 & 44.90 \\
        & 30 & 0.14 & 0.74 & 1189.89 &  0.15 & 0.81 & 310.52 &  0.12 & 0.6 & 713.90 & 0.10 & 0.73 & 70.65 \\
        & 40 & 0.14 & 0.71 & 1477.18 &  0.15 & 0.75 & 473.60 & 0.10 & 0.54 & 793.43 & 0.12 & 0.71 & 85.05 \\
        & 50  & 0.14 & 0.70 & 3432.57 & 0.14 & 0.75 & 575.12 & 0.10 & 0.53 & 848.85 & 0.10 & 0.67 & 129.48 \\
        \midrule
           \multirow{5}{*}{NMF-ner} & 10 & 0.08 & 0.68 & 1240.68 & 0.08 & 0.72 & 1059.28 & 0.08 & 0.71 & 649.24 & 0.10 & 0.73 & 298.26   \\
        & 20  & 0.07 & 0.58 & 1337.76 & 0.10 & 0.66 & 1158.13 & 0.10 & 0.72 & 698.68 & 0.11 & 0.67 & 314.34   \\
        & 30 & 0.09 & 0.57 & 1384.16 & 0.10 & 0.62 & 1213.91 &  0.11 & 0.6 & 713.90 & 0.0 & 0.55 & 338.26  \\
        & 40 & 0.08 & 0.48 & 1476.83 & 0.09 & 0.52 & 1568.97 &  0.10 & 0.54 & 793.43 & 0.10 & 0.52 &  342.19  \\
        & 50 & 0.08 & 0.48 & 6243.38 & 0.09 & 0.53 & 1660.62 & 0.10 & 0.53 & 848.86 & 0.09 & 0.47 & 387.96  \\
        \midrule               
        \multicolumn{14}{c}{\textbf{Neural-based Models - Default parameters}} \\
        \midrule        
             
        \multirow{5}{*}{BERTopic gte-base-en}
        & 10 & \cellcolor{lightyellow}\textbf{0.15} & 0.86 & \cellcolor{lightyellow}\textbf{110.89} & \cellcolor{lightyellow}\textbf{0.12} & \cellcolor{lightyellow}\textbf{0.90} & 255.98 & \cellcolor{lightyellow}\textbf{0.15} & \cellcolor{lightyellow}\textbf{0.92} & \cellcolor{lightyellow}\textbf{80.00} & 0.15 & 0.88 & \cellcolor{lightyellow}\textbf{28.97}
        \\
        & 20 & \cellcolor{lightyellow}\textbf{0.14} & \cellcolor{lightyellow}\textbf{0.88} & \cellcolor{lightyellow}\textbf{79.56} & 0.14 & \cellcolor{lightyellow}\textbf{0.83} & 291.60 & 0.13 & 0.82 & \cellcolor{lightyellow}\textbf{49.31} & \cellcolor{lightyellow}\textbf{0.16} & 0.79 & \cellcolor{lightyellow}\textbf{31.30}
        \\
        & 30 & \cellcolor{lightyellow}\textbf{0.15} & \cellcolor{lightyellow}\textbf{0.84} & \cellcolor{lightyellow}\textbf{85.04} & \cellcolor{lightyellow}\textbf{0.13} & 0.78 & 176.11 & 0.14 & \cellcolor{lightyellow}\textbf{0.85} & \cellcolor{lightyellow}\textbf{52.29} & 
         \cellcolor{lightyellow}\textbf{0.18} & \cellcolor{lightyellow}\textbf{0.79} & \cellcolor{lightyellow}\textbf{30.63}
        \\
        
        & 40 & \cellcolor{lightyellow}\textbf{0.16} & \cellcolor{lightyellow}\textbf{0.79} & \cellcolor{lightyellow}\textbf{84.35} & \cellcolor{lightyellow}\textbf{0.14} & 0.77 & 168.87 & 0.14 & \cellcolor{lightyellow}\textbf{0.79} & \cellcolor{lightyellow}\textbf{56.03} & \cellcolor{lightyellow}\textbf{0.18} & \cellcolor{lightyellow}\textbf{0.77} & \cellcolor{lightyellow}\textbf{26.76}
        \\
        
        & 50 & \cellcolor{lightyellow}\textbf{0.16} & \cellcolor{lightyellow}\textbf{0.77} & \cellcolor{lightyellow}\textbf{81.82} & \cellcolor{lightyellow}\textbf{0.15} & \cellcolor{lightyellow}\textbf{0.79} & 235.22 & \cellcolor{lightyellow}\textbf{0.16} & \cellcolor{lightyellow}\textbf{0.81} & \cellcolor{lightyellow}\textbf{52.40} & \cellcolor{lightyellow}\textbf{0.18}  & \cellcolor{lightyellow}\textbf{0.75} & \cellcolor{lightyellow}\textbf{31.15}        \\ 
        \midrule
        \multirow{5}{*}{BERTopic-jina}
        & 10 & 0.15 & \cellcolor{lightyellow}\textbf{0.88} & \cellcolor{lightyellow}\textbf{129.02} & 0.11 & 0.89 & 129.35 & \cellcolor{lightyellow}\textbf{0.16} & 0.90 & 92.19 & \cellcolor{lightyellow}\textbf{0.16} & \cellcolor{lightyellow}\textbf{0.93} & 33.09 
        \\
        & 20 & 0.14 & 0.82 & 107.97 & \cellcolor{lightyellow}\textbf{0.15} & 0.83 & \cellcolor{lightyellow}\textbf{99.73} & \cellcolor{lightyellow}\textbf{0.15} & \cellcolor{lightyellow}\textbf{0.87} & 68.43 & 0.16 & \cellcolor{lightyellow}\textbf{0.85} & 47.54 
        \\
        & 30 & 0.14 & 0.78 & 93.71 & 0.13 & \cellcolor{lightyellow}\textbf{0.79} & \cellcolor{lightyellow}\textbf{101.83} & 0.12 & 0.81 & 70.21 & 0.15 & 0.77 & 51.04 
        \\
        
        & 40 & 0.13 & 0.75 & 105.74 & \cellcolor{lightyellow}\textbf{0.15} & \cellcolor{lightyellow}\textbf{0.78} & \cellcolor{lightyellow}\textbf{102.76} & \cellcolor{lightyellow}\textbf{0.15} & 0.78 & 64.75 & 0.16 & 0.77 & 34.18
        \\
        
        & 50  & 0.14 & 0.73 & 105.47 & 0.15 & 0.77  & \cellcolor{lightyellow}\textbf{100.09} & 0.15 & 0.79 & 69.87 & 0.16 & 0.75 & 42.47 \\    

        \midrule

        \multicolumn{14}{c}{\textbf{Neural-based Models - Hypertuned parameters}} \\
        \midrule        
             
        \multirow{5}{*}{BERTopic gte-base-en}
        & 10 & \cellcolor{lightblue}\textbf{0.16} & \cellcolor{lightblue}\textbf{0.93} & \cellcolor{lightgreen}\textbf{33.09} & 0.10 & 0.88 & \cellcolor{lightgreen}\textbf{126.68} & \cellcolor{lightblue}\textbf{0.17} & 0.91   & \cellcolor{lightgreen}\textbf{74.30}  & 0.15 & 0.88 & \cellcolor{lightgreen}\textbf{44.63}
        \\
        & 20 & \cellcolor{lightblue}\textbf{0.16} & 0.85 & \cellcolor{lightgreen}\textbf{47.54} & 0.13 & 0.77 & \cellcolor{lightgreen}\textbf{70.59} & \cellcolor{lightblue}\textbf{0.16} & \cellcolor{lightblue}\textbf{0.86}  & \cellcolor{lightgreen}\textbf{44.77}  & \cellcolor{lightblue}\textbf{0.18} & 0.78 & \cellcolor{lightgreen}\textbf{18.45} 
        \\
        & 30  & 0.15 & 0.77 & \cellcolor{lightgreen}\textbf{51.04} & \cellcolor{lightblue}\textbf{0.14} & 0.74 & \cellcolor{lightgreen}\textbf{75.78} & \cellcolor{lightblue}\textbf{0.15} & 0.79 & \cellcolor{lightgreen}\textbf{46.72}  & \cellcolor{lightblue}\textbf{0.19} & 0.72 & \cellcolor{lightgreen}\textbf{19.37} 
        \\
        
        & 40 & \cellcolor{lightblue}\textbf{0.17} & 0.77 & \cellcolor{lightgreen}\textbf{34.19} & \cellcolor{lightblue}\textbf{0.15} & 0.73 & \cellcolor{lightgreen}\textbf{73.97} & \cellcolor{lightblue}\textbf{0.18} & 0.79 & \cellcolor{lightgreen}\textbf{46.55} & 0.17 & 0.67 & \cellcolor{lightgreen}\textbf{16.36} 
        \\
        
        & 50 &  0.16 & 0.75 & \cellcolor{lightgreen}\textbf{42.47} &  \cellcolor{lightblue}\textbf{0.16} & 0.75 & \cellcolor{lightgreen}\textbf{78.51} & \cellcolor{lightblue}\textbf{0.17} & 0.77   & \cellcolor{lightgreen}\textbf{44.30} &  0.18 & 0.67 & \cellcolor{lightgreen}\textbf{19.12} \\ 
        \midrule
        \multirow{5}{*}{BERTopic-jina}
        & 10 & 0.13 & 0.80 & \cellcolor{lightgreen}\textbf{106.09} & 0.08 & 0.81 & \cellcolor{lightgreen}\textbf{125.39} & 0.15 & \cellcolor{lightblue}\textbf{0.92} & \cellcolor{lightgreen}\textbf{80.83} & \cellcolor{lightblue}\textbf{0.18} & 0.84 & \cellcolor{lightgreen}\textbf{44.78} 
        \\
        & 20 & 0.14 & 0.76 & \cellcolor{lightgreen}\textbf{98.40} & 0.12 & 0.76 & \cellcolor{lightgreen}\textbf{96.33} & 0.15 & 0.83 & \cellcolor{lightgreen}\textbf{63.47} & \cellcolor{lightblue}\textbf{0.19} & 0.78   & \cellcolor{lightgreen}\textbf{41.38} 
        \\
        & 30 & 0.14 & 0.74 & \cellcolor{lightgreen}\textbf{87.85} & 0.14 & 0.75 & \cellcolor{lightgreen}\textbf{92.59} & 0.14 & 0.79 & \cellcolor{lightgreen}\textbf{63.74} & \cellcolor{lightblue}\textbf{0.18} & 0.70 & \cellcolor{lightgreen}\textbf{40.34} 
        \\
        
        & 40 & 0.15 & 0.73 & \cellcolor{lightgreen}\textbf{96.34} & 0.16 & 0.74 & \cellcolor{lightgreen}\textbf{95.80} & 0.16 & \cellcolor{lightblue}\textbf{0.80} & \cellcolor{lightgreen}\textbf{62.22} & \cellcolor{lightblue}\textbf{0.17} & 0.66 & \cellcolor{lightgreen}\textbf{43.34}
        \\
        
        & 50  & 0.15 & 0.71 & \cellcolor{lightgreen}\textbf{95.22} & 0.16 & 0.75  & 106.60  & 0.17 & 0.76 & \cellcolor{lightgreen}\textbf{56.93} & \cellcolor{lightblue}\textbf{0.17} & 0.65  & \cellcolor{lightgreen}\textbf{36.88}  \\           
        \bottomrule        
\end{tabular}
\caption{Quantitative comparison of various methods across topic counts (\#T). Highlighted values indicate top scores per \#T.}
\label{tab:quantitative results}
\end{table*}

 
\subsection{\textit{Topic-Modeling Evaluation Framework}}
We evaluate the quality of discovered topics using a comprehensive framework that combines quantitative metrics for static BERTopic modeling and qualitative analysis for both static and dynamic models. This multi-faceted approach ensures reliable, consistent, and comprehensible results, which are especially valuable for analyzing large-scale textual data. Dynamic modeling further enables longitudinal analysis, allowing researchers to track topic evolution over time, an essential feature for historical, social, and trend studies. The structured evaluation framework also promotes reproducibility, facilitating replication and validation across similar datasets.

\textit{Quantitative Assessment:} We employ two key quantitative metrics to assess the quality of extracted topics systematically. \textit{\textcolor{black}{Topic Coherence}} measures the semantic similarity of words within a topic, ensuring that each topic consists of closely related terms. A higher coherence score indicates that the words within a topic are contextually meaningful. \textit{\textcolor{black}{Topic Diversity}}, on the other hand, evaluates the distinctiveness of topics across the dataset, reducing redundancy and ensuring comprehensive theme coverage. Maintaining a balance between coherence and diversity enhances the understanding of topics. 
 We compare the results of BERTopic with two classical topic-modeling techniques, LDA and NMF, to validate its performance and effectiveness. Additionally, we analyze the \textit{Topic size distribution} to examine the balance of topic representation across documents, identifying potential biases where certain topics dominate while others remain underrepresented. This helps understand the overall structure of the dataset. 

\textit{Qualitative Assessment:}  
In addition to quantitative metrics, we conduct a thorough qualitative analysis to assess the human understandability and relevance of the topics identified by BERTopic. \textit{\textcolor{black}{Static Topic Modeling:}} For static BERTopic modeling, we begin with a review of a sample of documents from each topic to verify that they accurately reflect the assigned theme. We also inspect the key terms defining each topic to ensure they align semantically and contextually with the dataset. To further validate the results, domain experts reviewed the selected topics to assess their real-world relevance and human understandability, ensuring that the themes captured are meaningful and useful for the research. Since static BERTopic modeling generates topics based on the entire dataset, this comprehensive qualitative evaluation helps confirm that the extracted topics are coherent, distinct, and aligned with the overall content of the data to ensure the reliability of the model’s findings. \textit{\textcolor{black}{Dynamic Topic Modeling and Evolution Analysis:}} Dynamic BERTopic enables the visualization of topic evolution over time, providing insights into how themes emerge, change, and disappear across different periods. This temporal analysis is especially useful for studying historical trends, shifts in discourse, and long-term patterns in large datasets. By examining these dynamics, we gain deeper insights into how societal or thematic changes unfold over time.
\renewcommand{\arraystretch}{1}

\begin{table*}[ht!]
\centering
\caption{Qualitative comparison of topic keywords across two embedding models (BERTopic - GTE and Jina), highlighting common themes and differences.}
\label{tab:qualitative analysis}

\begin{tabularx}{\textwidth}{|p{2.22cm}|p{3cm}|p{3cm}|X|}
\hline
\textbf{Theme} & \textbf{GTE Keywords} & \textbf{Jina Keywords} & \textbf{Observations} \\
\hline

\textbf{Nuclear Weapons \& Cold War Politics} & \texttt{soviet, american, president, united, nuclear, states, weapon, new, missile, party} & \texttt{oil, country, nuclear, energy, minister, united, president, Israel, states, new} & \textbf{Common Theme}: Both involve geopolitics and nuclear power, though GTE focuses on nuclear weapons and military politics (e.g., "missile," "weapon"). \newline \textbf{Differences}: GTE centers around the Cold War arms race, while Jina involves broader international political dynamics, including oil and Israel. \\
\hline

\textbf{Nuclear Accidents \& Chernobyl} & \texttt{nuclear, reactor, plant, uranium, power, accident, Chernobyl, fuel, energy, country} & \texttt{chernobyl, accident, reactor, plant, power, nuclear, soviet, disaster, radioactivity, accord} & \textbf{Common Theme}: Both sets reference nuclear accidents, particularly Chernobyl, and the nuclear plant. \newline \textbf{Differences}: GTE is more focused on the technical aspects of reactors and energy, while Jina highlights the disaster and the global political implications ("Soviet," "radioactivity"). \\
\hline 
\textbf{Environmental \& Ecological Concerns} & \texttt{seveso, dioxin, icmesa, givaudan, roche, hoffmannla, trial, monza, factory, accident'} & \texttt{seveso, dioxin, icmesa,givaudan, rhine, montlouis, chemical, factory, barrel} & \textbf{Common Theme}: Both sets focus on the Seveso disaster and its environmental impact, particularly related to chemical pollution and industrial accidents. \newline \textbf{Differences}: GTE focuses more on legal aspects (e.g., "trial"), while Jina emphasizes broader environmental and geographical context (e.g., "Rhine," "barrel"). \\
\hline

\textbf{International Political Relations} & \texttt{china, Chinese, beijing, nuclear, visit, soviet, american, united, president, chinas} & \texttt{chinese, china, beijing, vietnam, american, north, mao, Vietnamese, south, president} & \textbf{Common Theme}: Both sets involve international politics, focusing on countries like China and the United States. \newline \textbf{Differences}: GTE mentions nuclear visits and relations between superpowers, while Jina refers more to historical political dynamics in China and Vietnam. \\
\hline 

\textbf{Disarmament and Peace Initiatives} & \texttt{church, bishop, pope, paul, john, peace, churches, wcc, nuclear, catholic} & \texttt{disarmament, peace, weapon, nuclear, treaty, conference, church, world, war, geneva} & \textbf{Common Theme}: Both sets involve peace and disarmament initiatives, with a focus on nuclear disarmament. \newline \textbf{Differences}: The first set focuses on religious and church-related peace efforts, while the second set focuses more on political and global disarmament efforts, such as treaties and conferences. \\
\hline 
\end{tabularx}
\end{table*}


\subsection{\textit{Static Topic-Modeling Methods Performance}}
This section presents both the quantitative and qualitative evaluation results of static topic modeling methods.

\subsubsection{\textit{Quantitative Results}} Table~\ref{tab:quantitative results} presents the comparative performance of various static topic modeling approaches, including LDA and its variants, NMF and its variants, as well as BERTopic with different embedding models. The best scores for the BERTopic variants (GTE and Jina) in the default setting are highlighted in \textit{yellow} for internal comparison, while in the hypertuned setting, scores are highlighted in \textit{blue} and \textit{green} and compared separately for each embedding model.

Since classical topic models such as LDA and NMF require the number of topics to be specified in advance, we tested a range of topic counts (10, 20, 30, 40, and 50) to evaluate their performance across different configurations. In contrast, BERTopic determines the number of topics dynamically. To enable a fair comparison with classical models, we initially ran BERTopic in its default configuration and subsequently reduced the number of topics to match the specified counts used in the classical models. This approach allowed us to directly compare performance metrics across methods while accounting for differences in topic generation strategies. 

Overall, BERTopic configured with the \textit{gte-base embedding model} consistently outperformed all other methods across our key evaluation metrics, namely topic coherence, topic diversity, and computational efficiency. The variant using Jina embeddings demonstrated comparable performance, suggesting that both models are well-suited for capturing semantically rich and coherent topics in large-scale textual datasets, such as historical newspaper archives. These results reinforce the idea that embedding quality and sequence-handling capabilities are more important than model size alone for this task. Notably, although Jina embeddings offer a more compact representation, their performance was on par with the larger GTE-based model. This highlights that, for many applications, particularly in topic modeling, the choice of the embedding model should be driven by the ability to effectively capture semantic meaning and handle long sequences, rather than by the sheer size of the model.

Classical models such as LDA and NMF, particularly those using TF-IDF representations, exhibited limited performance and often produced only a small number of topics, even when initialized with a predefined topic count. While augmenting these models with Named Entity Recognition (NER) offered modest improvements in topic understandability, it also introduced substantial computational overhead. The addition of NER increased processing time, reducing the overall efficiency of these models. Despite these enhancements, the performance of the classical models remained significantly lower than that of BERTopic. This suggests that while NER can help improve the specificity of topic extraction, it is not sufficient to overcome the inherent limitations of classical models compared to more contextually aware approaches like BERTopic.

We also experimented with hyperparameter optimization for BERTopic to assess potential performance gains. While hyperparameter tuning reduced computation time during inference and slightly improved topic coherence in some cases, the overall topic quality remained largely unchanged compared to the default configuration. Moreover, the optimization process itself was computationally expensive, taking over 24 hours to complete on the GPU infrastructure, and, in some cases, led to overfitting, producing fewer, overly specific topics. These findings indicate that extensive hyperparameter tuning may not be necessary for most applications, particularly when computational resources are limited, as the default BERTopic configuration already offers strong performance across diverse datasets.

\begin{figure*}[ht!]
\centering

\includegraphics[width=0.9\textwidth]{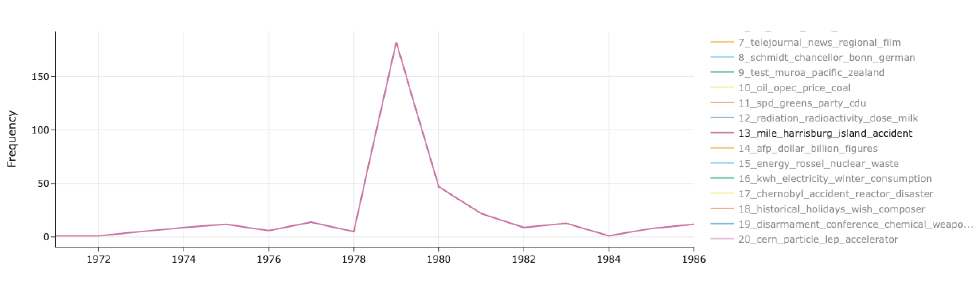}
\\
\textbf{(a)} Topic dynamics from 1971 to 1986 showing significant spikes in public discourse during the Harrisburg (1979) and Chernobyl (1986) nuclear incidents.

\vspace{1em}

\includegraphics[width=0.9\textwidth]{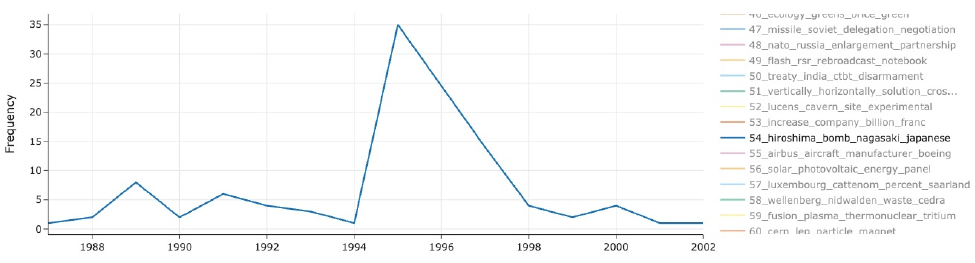}
\\
\textbf{(b)} Topic spike in 1995 reflecting the global discourse surrounding the 50th anniversary of the Hiroshima bombing, with a focus on nuclear disarmament and peace.

\vspace{1em}

\includegraphics[width=0.9\textwidth]{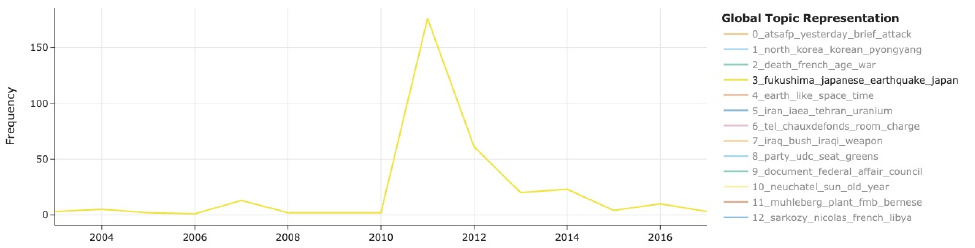}
\\
\textbf{(c)} Topic dynamics from 2003 to 2018, highlighting a sharp rise in attention during the Fukushima earthquake and nuclear crisis in 2011.

\caption{Topic evolution visualizations for key historical events: Harrisburg \& Chernobyl, Hiroshima 50th anniversary, and Fukushima disaster.}
\label{fig:topic_evolution_combined}
\end{figure*}

\subsubsection{\textit{Qualitative Results}} In addition to the quantitative evaluation, a qualitative review was conducted to assess the clarity and practical relevance of the generated topics. Table~\ref{tab:qualitative analysis} presents a sample of the qualitative evaluation of five topics from the best-performing models, selected based on their quantitative scores. Both BERTopic with GTE embeddings and Jina demonstrated strong capabilities in clustering topic-related keywords across a variety of themes. However, the models exhibited distinct differences in their approach to topic extraction. GTE predominantly emphasized technical, legal, and political dimensions of topics, while Jina focused more on geopolitical and environmental concerns. These variations suggest that both models are adaptable to different domains, with each excelling in capturing specific thematic aspects depending on the dataset. The selection of the embedding model should thus depend on the particular domain, the types of topics to be extracted, and the required level of specificity or generalization.

In contrast, classical methods like LDA and NMF generated fewer topics, often producing less coherent and less relevant themes that did not align well with the actual content of the documents. In many cases, these methods generated generic topics with limited clarity, requiring significant post-processing to make them meaningful. On the other hand, BERTopic, through its use of contextual embeddings, was able to extract more nuanced topics that captured subtle shifts in language and thematic emphasis, crucial for understanding complex datasets such as historical records.

\textbf{\textit{Findings.}} Our results confirm the superiority of embedding-based topic models for large-scale, heterogeneous corpora. Sentence embeddings from models like GTE, Jina, or any other model significantly enhance semantic understanding and topic coherence without requiring extensive hyperparameter tuning. In resource-constrained environments, BERTopic, with its default settings, provides an efficient and robust solution, outperforming classical approaches even without extensive optimization. These findings align with broader trends in the literature, supporting the effectiveness of neural embedding models in real-world topic-modeling applications \citep{bianchi2021pre,dieng-etal-2020-topic}. Additionally, our experiments suggest that, in many scenarios—especially when dealing with large, diverse datasets—the default configurations of BERTopic are not only adequate but often optimal. This reinforces the argument that simplicity in hyperparameter selection can be preferable to complexity, particularly when time and computational resources are limited.

Moreover, the qualitative findings reinforce the conclusions from the quantitative evaluation that BERTopic, particularly when using pre-configured embeddings, offers a more effective and efficient solution. Overall, the qualitative results further support the broader trends in the literature, affirming that embedding-based models, such as BERTopic, provide a more effective and scalable solution for topic modeling, especially when dealing with large, complex, and domain-specific datasets. The ability of these models to generate meaningful, domain-specific topics with minimal tuning makes them highly suitable for real-world applications in fields like historical data analysis, political discourse analysis, and multi-domain text mining.

\subsection{Dynamic Topic-Modeling Performance}

Based on our qualitative evaluation, BERTopic consistently outperformed classical models in terms of topic clarity and relevance. Moreover, the choice of the embedding model (e.g., GTE or Jina) can be tailored to the user's domain interests, offering flexibility depending on the thematic focus---e.g., political, legal, environmental, or geopolitical. Building on these findings, we incorporated temporal information (i.e., publication year) to analyze how topics evolve. We prepared the dataset for dynamic topic modeling and applied BERTopic’s temporal modeling capabilities to achieve this. Unlike in the qualitative evaluation, where we reduced the number of topics to predefined counts (10, 20, 30, 40, 50) for comparative purposes, we now retained the original number of topics automatically identified by BERTopic. This allowed for a more natural representation of topic dynamics over time, as determined by the model’s clustering process. Since the dynamic topic visualizations generated using Plotly are best explored in an interactive environment, in this paper, we present selected examples to illustrate how topic evolution can be captured effectively. We focus on historically significant and widely recognizable events, such as the Harrisburg accident, the Chernobyl disaster, the Hiroshima bombing and its related documentary coverage, and the Fukushima earthquake. The corresponding topic evolution plots immediately highlight temporal spikes (see Fig.~\ref{fig:topic_evolution_combined}.), in discussion around these events, providing a visual narrative of their impact within the dataset. The x-axis of the plots represents the frequency of topics, reflecting their prominence across different time periods (or document frequency). 

We highlight a few historically significant events that are well-represented in our dataset. The Three Mile Island nuclear accident, which occurred in Harrisburg, Pennsylvania, in 1979, marked one of the most serious nuclear incidents in U.S. history. Given its scale and media attention at the time, we observe a distinct surge in topic activity around this period, reflecting widespread public concern and extensive journalistic coverage. Similarly, the Chernobyl disaster in 1986 -- widely regarded as the most catastrophic nuclear accident -- triggered a global response, and its long-term environmental, political, and health implications are clearly visible in the temporal trajectory of associated topics. The model captures not only the immediate spike in discussion but also the recurring attention during subsequent anniversaries and related events. The Hiroshima bombing in 1945, while predating the majority of digitized news archives, saw renewed attention in 1995 following the release of a widely discussed documentary marking the 50th anniversary of the event. This reemergence of discourse around Hiroshima is evident in our dynamic topic plots, suggesting how cultural artifacts such as documentaries can reignite public and media interest in historical events. Finally, the 2011 Fukushima earthquake and nuclear crisis generated another marked spike in topic activity. The event's compound nature -- a powerful earthquake followed by a tsunami and nuclear failure -- led to global coverage and policy debates, which are prominently reflected in the topic evolution patterns.

These case studies demonstrate how dynamic topic modeling, particularly when integrated with temporal metadata, enables researchers to detect and understand the historical patterns in large-scale textual corpora. The resulting visualizations provide an accessible and data-driven narrative of how public discourse evolves around major events, offering a powerful tool for both computational analysis and historical inquiry.

\section{Conclusion and Future Work}
In this work, we demonstrate that BERTopic outperforms classical models like LDA, NMF, and their variants in capturing complex patterns and thematic shifts within large, unstructured textual collections, such as historical newspaper archives. While larger models may offer better representations, the choice of the embedding model in BERTopic should be guided by user requirements, focusing on pretrained data and the specific area for topic extraction, rather than solely on model size. For instance, models like GTE and Jina, though similar in size, exhibit differences in topic extraction due to variations in their pre-trained data. Hyperparameter optimization is not always necessary, particularly when resources are constrained. Both static and dynamic topic modeling with BERTopic offer valuable insights, and the future improvements discussed in the limitations will further enhance human understandability and usability for large-scale text analysis.

While BERTopic excels in modeling both static and dynamic topic structures, there are opportunities to improve topic clarity and usability, especially for large, temporally distributed datasets like historical newspaper archives. A key limitation of static modeling is the assignment of each document to a single topic, which can obscure overlapping themes. To address this, we propose directly incorporating soft clustering techniques, such as probabilistic topic distributions, into BERTopic. This would allow documents to be assigned to multiple topics, providing a more nuanced representation.  Additionally, static models often produce redundant or semantically similar topics, which can undermine clarity. We recommend user-guided or semi-automated merging of related topics based on embedding similarity or shared keywords. For dynamic topic modeling, these refinements are crucial to prevent fragmented topics from distorting topic evolution visualizations. Enhancing these visualizations with collapsible groupings and links to representative documents will aid users in understanding emerging trends and shifts, leading to a more coherent and human-understandable topic-modeling framework for complex textual collections.

\section{Acknowledgements}
We thank the Impresso team for providing the document collection used in this study, based on their project \href{https://impresso-project.ch}{Impresso - Media Monitoring of the Past II} , funded by the Swiss National Science Foundation (SNSF 213585) and the Luxembourg National Research Fund (17498891). 

\section{Funding}
This work was supported by Doctoral Training Unit - Deep Data Science for History (D4H)\footnote{\url{https://d4h.uni.lu/}}, a collaboration between the Luxembourg Centre for Contemporary and Digital History (C$^2$DH) and the Department of Computer Science (DCS), both at the University of Luxembourg, aimed at fostering interdisciplinary research between Historians and Computer Scientists.

\section{Data Availability} The code and data will be publicly available on GitHub upon acceptance. \href{https://github.com/KeerthanaMurugaraj/Decoding-History-with-Neural-Topic-Modeling-for-Analyzing-Unstructured-Newspaper-Archives}{Link}

\section{Author contributions statement}
K.M. led the conceptualization, methodology, data analysis, software implementation, and writing of the manuscript. S.L. contributed to the methodology and participated in manuscript review and editing. M.D. contributed resources and participated in the evaluation of the results. M.T. supervised the project and contributed to its administration, as well as the review and editing of the manuscript.

\section{Additional information}
Correspondence and requests for materials should be addressed to K.M
\section{{Competing interests:}}
The author(s) declare no competing interests.

\begin{appendices}

\end{appendices}

\bibliographystyle{abbrvnat}
\bibliography{reference.bib}

@article{10.1093/llc/fqab099,
    author = {Huang, Shi-Yun and Wu, Shang-Yun and Chen, You-Jun and Tsai, Richard Tzong-Han and Fan, I-Chun},
    title = {Climate event classification based on historical meteorological records and its presentation on a Spatio-Temporal research platform},
    journal = {Digital Scholarship in the Humanities},
    volume = {37},
    number = {4},
    pages = {1022-1032},
    year = {2022},
    month = {01},
    issn = {2055-7671},
    doi = {10.1093/llc/fqab099},
    url = {https://doi.org/10.1093/llc/fqab099},
    eprint = {https://academic.oup.com/dsh/article-pdf/37/4/1022/46607958/fqab099.pdf},
}

@inproceedings{Ehrmann2020HistoricalNC,
  title={{Historical Newspaper Content Mining: Revisiting the Impresso Project's Challenges in Text and Image Processing, Design and Historical Scholarship}},
  author={Maud Ehrmann and Estelle Bunout and S. Clematide and Marten D{\"u}ring and Andreas Fickers and Roman Kalyakin and F. Kaplan and Matteo Romanello and Paul Schroeder and Philipp Str{\"o}bel and Thijs van Beek and Martin Volk and Lars Wieneke},
  booktitle={{Digital Humanities Conference}},
  year={2020},
  url={https://api.semanticscholar.org/CorpusID:220667683}
}

@article{10.1093/llc/fqaf024,
    author = {Tsai, Richard Tzong-Han and Liu, Yu-Sin and Chen, You-Jun and Hsieh, Hsin Yi and Chan, Ya-Chi},
    title = {Digital humanities approach to analyzing the roles and military power of Supreme Commanders and Grand Coordinators in the Ming Dynasty: a computational analysis of Ming Shilu},
    journal = {Digital Scholarship in the Humanities},
    pages = {fqaf024},
    year = {2025},
    month = {05},
    
    issn = {2055-7671},
    doi = {10.1093/llc/fqaf024},
    url = {https://doi.org/10.1093/llc/fqaf024},
    eprint = {https://academic.oup.com/dsh/advance-article-pdf/doi/10.1093/llc/fqaf024/63327229/fqaf024.pdf},
}

@article{10.1093/llc/fqaf038,
    author = {Vorobeva, Viktoriia and Bonch-Osmolovskaya, Anastasia and Kryukov, Artem and Podriadchikova, Maria},
    title = {Distant reading of Soviet diaries},
    journal = {Digital Scholarship in the Humanities},
    pages = {fqaf038},
    year = {2025},
    month = {05},
    issn = {2055-7671},
    doi = {10.1093/llc/fqaf038},
    url = {https://doi.org/10.1093/llc/fqaf038},
    eprint = {https://academic.oup.com/dsh/advance-article-pdf/doi/10.1093/llc/fqaf038/63300446/fqaf038.pdf},
}

@article{10.1093/llc/fqaf037,
    author = {Cvrček, Václav and Berrocal, Martina},
    title = {Sibling-texts keyword analysis: exploring topic and register keywords},
    journal = {Digital Scholarship in the Humanities},
    pages = {fqaf037},
    year = {2025},
    month = {05},
    issn = {2055-7671},
    doi = {10.1093/llc/fqaf037},
    url = {https://doi.org/10.1093/llc/fqaf037},
    eprint = {https://academic.oup.com/dsh/advance-article-pdf/doi/10.1093/llc/fqaf037/63217235/fqaf037.pdf},
}

@article{10.1093/llc/fqaf036,
    author = {Yamashita, Taisei and Uchida, Atsuhiko},
    title = {Historical analysis using topic modelling: insights from Khevenhüller’s diary 1742–76},
    journal = {Digital Scholarship in the Humanities},
    pages = {fqaf036},
    year = {2025},
    month = {05},
    issn = {2055-7671},
    doi = {10.1093/llc/fqaf036},
    url = {https://doi.org/10.1093/llc/fqaf036},
    eprint = {https://academic.oup.com/dsh/advance-article-pdf/doi/10.1093/llc/fqaf036/63056691/fqaf036.pdf},
}

@article{10.1093/llc/fqaf025,
    author = {Sun, Shaodan and Qin, Xugong},
    title = {Fine-grained extraction of geospatial and temporal information from Chinese historical newspapers},
    journal = {Digital Scholarship in the Humanities},
    pages = {fqaf025},
    year = {2025},
    month = {03},
    issn = {2055-7671},
    doi = {10.1093/llc/fqaf025},
    url = {https://doi.org/10.1093/llc/fqaf025},
    eprint = {https://academic.oup.com/dsh/advance-article-pdf/doi/10.1093/llc/fqaf025/62749321/fqaf025.pdf},
}

@inproceedings{bianchi2021pre,
  title={Pre-training is a Hot Topic: Contextualized Document Embeddings Improve Topic Coherence},
  author={Bianchi, Federico and Terragni, Silvia and Hovy, Dirk},
  booktitle={Proceedings of the 59th Annual Meeting of the Association for Computational Linguistics},
  pages={759--766},
  year={2021}
}

@article{dieng-etal-2020-topic,
    title = "Topic Modeling in Embedding Spaces",
    author = "Dieng, Adji B.  and
      Ruiz, Francisco J. R.  and
      Blei, David M.",
    editor = "Johnson, Mark  and
      Roark, Brian  and
      Nenkova, Ani",
    journal = "Transactions of the Association for Computational Linguistics",
    volume = "8",
    year = "2020",
    address = "Cambridge, MA",
    publisher = "MIT Press",
    url = "https://aclanthology.org/2020.tacl-1.29/",
    doi = "10.1162/tacl_a_00325",
    pages = "439--453",
}

@inproceedings{Reimers2019SentenceBERTSE,
  title={Sentence-BERT: Sentence Embeddings using Siamese BERT-Networks},
  author={Nils Reimers and Iryna Gurevych},
  booktitle={Conference on Empirical Methods in Natural Language Processing},
  year={2019},
  url={https://api.semanticscholar.org/CorpusID:201646309}
}

@inproceedings{Vaswani2017AttentionIA,
  title={Attention is All you Need},
  author={Ashish Vaswani and Noam M. Shazeer and Niki Parmar and Jakob Uszkoreit and Llion Jones and Aidan N. Gomez and Lukasz Kaiser and Illia Polosukhin},
  booktitle={Neural Information Processing Systems},
  year={2017},
  url={https://api.semanticscholar.org/CorpusID:13756489}
}

@inproceedings{devlin-etal-2019-bert,
    title = "{BERT}: Pre-training of Deep Bidirectional Transformers for Language Understanding",
    author = "Devlin, Jacob  and
      Chang, Ming-Wei  and
      Lee, Kenton  and
      Toutanova, Kristina",
    editor = "Burstein, Jill  and
      Doran, Christy  and
      Solorio, Thamar",
    booktitle = "Proceedings of the 2019 Conference of the North {A}merican Chapter of the Association for Computational Linguistics: Human Language Technologies, Volume 1 (Long and Short Papers)",
    month = jun,
    year = "2019",
    address = "Minneapolis, Minnesota",
    publisher = "Association for Computational Linguistics",
    url = "https://aclanthology.org/N19-1423/",
    doi = "10.18653/v1/N19-1423",
    pages = "4171--4186",
}

@misc{McInnes2018UMAPUM,
      title={{UMAP: Uniform Manifold Approximation and Projection for Dimension Reduction}}, 
      author={Leland McInnes and John Healy and James Melville},
      year={2020},
      eprint={1802.03426},
      archivePrefix={arXiv},
      primaryClass={stat.ML},
       
}

@article{McInnes2017hdbscanHD,
  title={{hdbscan: Hierarchical density-based clustering}},
  author={Leland McInnes and John Healy and S. Astels},
  journal={Journal of Open Source Software.},
  year={2017},
  volume={2},
  pages={205},
 }

@inproceedings{10.1145/544220.544222,
author = {Tibbo, Helen R.},
title = {Primarily history: historians and the search for primary source materials},
year = {2002},
isbn = {1581135130},
publisher = {Association for Computing Machinery},
address = {New York, NY, USA},
doi = {10.1145/544220.544222},
booktitle = {Proceedings of the 2nd ACM/IEEE-CS Joint Conference on Digital Libraries},
pages = {1–10},
numpages = {10},
location = {Portland, Oregon, USA},
series = {JCDL '02}
}

@article{Grootendorst2022BERTopicNT,
  title={BERTopic: Neural topic modeling with a class-based TF-IDF procedure},
  author={Maarten R. Grootendorst},
  journal={ArXiv},
  year={2022},
  volume={abs/2203.05794},
  }

@inproceedings{Bergstra2013HyperoptAP,
  title={Hyperopt: A Python Library for Optimizing the Hyperparameters of Machine Learning Algorithms},
  author={James Bergstra and Daniel Yamins and David D. Cox},
  booktitle={SciPy},
  year={2013},
 }

@inproceedings{NIPS2011_86e8f7ab,
 author = {Bergstra, James and Bardenet, R\'{e}mi and Bengio, Yoshua and K\'{e}gl, Bal\'{a}zs},
 booktitle = {Advances in Neural Information Processing Systems},
 editor = {J. Shawe-Taylor and R. Zemel and P. Bartlett and F. Pereira and K.Q. Weinberger},
 publisher = {Curran Associates, Inc.},
 title = {Algorithms for Hyper-Parameter Optimization},
  volume = {24},
 year = {2011}
}

@article{10.5555/944919.944937,
author = {Blei, David M. and Ng, Andrew Y. and Jordan, Michael I.},
title = {Latent dirichlet allocation},
year = {2003},
issue_date = {3/1/2003},
publisher = {JMLR.org},
volume = {3},
ISSN = {1532-4435},
journal = {J. Mach. Learn. Res.},
month = mar,
pages = "993–1022"
}

@article{Lee1999,
author={Lee, Daniel D.
and Seung, H. Sebastian},
title={Learning the parts of objects by non-negative matrix factorization},
journal={Nature},
year={1999},
month={Oct},
day={01},
volume={401},
number={6755},
pages={788-791},
issn={1476-4687},
doi={10.1038/44565},
}

@article{Deerwester1990IndexingBL,
  title={Indexing by Latent Semantic Analysis},
  author={Scott C. Deerwester and Susan T. Dumais and Thomas K. Landauer and George W. Furnas and Richard A. Harshman},
  journal= {J. Am. Soc. Inf. Sci.},
  year={1990},
  volume={41},
  pages={391-407},
 }

@article{article_Gillis,
author = {Gillis, Nicolas and Vavasis, Stephen},
year = {2012},
month = {08},
title = {Fast and Robust Recursive Algorithms for Separable Nonnegative Matrix Factorization},
volume = {36},
journal = {IEEE Transactions on Pattern Analysis and Machine Intelligence},
doi = {10.1109/TPAMI.2013.226}
}

@article{Kumar2012FastCH,
  title={Fast Conical Hull Algorithms for Near-separable Non-negative Matrix Factorization},
  author={Abhishek Kumar and Vikas Sindhwani and Prabhanjan Kambadur},
  journal={ArXiv},
  year={2012},
  volume={abs/1210.1190},
  }

@article{article_Gillis_Nicolas,
author = {Gillis, Nicolas},
year = {2013},
month = {10},
title = {Successive Nonnegative Projection Algorithm for Robust Nonnegative Blind Source Separation},
volume = {7},
journal = {SIAM Journal on Imaging Sciences [electronic only]},
doi = {10.1137/130946782}
}

@article{8666058,
  author={Chen, Yong and Wu, Junjie and Lin, Jianying and Liu, Rui and Zhang, Hui and Ye, Zhiwen},
  journal={IEEE Transactions on Knowledge and Data Engineering}, 
  title={Affinity Regularized Non-Negative Matrix Factorization for Lifelong Topic Modeling}, 
  year={2020},
  volume={32},
  number={7},
  pages={1249-1262},
  doi={10.1109/TKDE.2019.2904687}}

@article{10.1109/TPAMI.2016.2554555,
author = {Trigeorgis, George and Bousmalis, Konstantinos and Zafeiriou, Stefanos and Schuller, Bjorn W.},
title = {A Deep Matrix Factorization Method for Learning Attribute Representations},
year = {2017},
issue_date = {March 2017},
publisher = {IEEE Computer Society},
address = {USA},
volume = {39},
number = {3},
issn = {0162-8828},
doi = {10.1109/TPAMI.2016.2554555},
journal = {IEEE Trans. Pattern Anal. Mach. Intell.},
month = {mar},
pages = {417–429},
numpages = {13}
}

@article{10.1016/j.neucom.2022.10.002,
author = {Wang, Jianyu and Zhang, Xiao-Lei},
title = {Deep NMF topic modeling},
year = {2023},
issue_date = {Jan 2023},
publisher = {Elsevier Science Publishers B. V.},
address = {NLD},
volume = {515},
number = {C},
issn = {0925-2312},
doi = {10.1016/j.neucom.2022.10.002},
journal = {Neurocomput.},
month = {jan},
pages = {157–173},
numpages = {17}
}

@inproceedings{10.1145/312624.312649,
author = {Hofmann, Thomas},
title = {Probabilistic latent semantic indexing},
year = {1999},
isbn = {1581130961},
publisher = {Association for Computing Machinery},
address = {New York, NY, USA},
doi = {10.1145/312624.312649},
booktitle = {Proceedings of the 22nd Annual International ACM SIGIR Conference on Research and Development in Information Retrieval},
pages = {50–57},
numpages = {8},
location = {Berkeley, California, USA},
series = {SIGIR '99}
}

@inproceedings{10.5555/2073796.2073829,
author = {Hofmann, Thomas},
title = {Probabilistic latent semantic analysis},
year = {1999},
isbn = {1558606149},
publisher = {Morgan Kaufmann Publishers Inc.},
address = {San Francisco, CA, USA},
booktitle = {Proceedings of the Fifteenth Conference on Uncertainty in Artificial Intelligence},
pages = {289–296},
numpages = {8},
location = {Stockholm, Sweden},
series = {UAI'99}
}

@inproceedings{Hofmann1999ProbabilisticLS,
  title={Probabilistic Latent Semantic Analysis},
  author={Thomas Hofmann},
  booktitle={Conference on Uncertainty in Artificial Intelligence},
  year={1999},
 }

@article{10.1111/j.2517-6161.1977.tb01600.x,
    author = {Dempster, A. P. and Laird, N. M. and Rubin, D. B.},
    title = "{Maximum Likelihood from Incomplete Data Via the EM Algorithm}",
    journal = {Journal of the Royal Statistical Society: Series B (Methodological)},
    volume = {39},
    number = {1},
    pages = {1-22},
    year = {2018},
    month = {12},
    issn = {0035-9246},
    doi = {10.1111/j.2517-6161.1977.tb01600.x},
    }

@inproceedings{NIPS2005_9e82757e,
 author = {Lafferty, John and Blei, David},
 booktitle = {Advances in Neural Information Processing Systems},
 editor = {Y. Weiss and B. Sch\"{o}lkopf and J. Platt},
 publisher = {MIT Press},
 title = {Correlated Topic Models},
  volume = {18},
 year = {2005}
}

@inproceedings{10.1145/1143844.1143859,
author = {Blei, David M. and Lafferty, John D.},
title = {Dynamic topic models},
year = {2006},
isbn = {1595933832},
publisher = {Association for Computing Machinery},
address = {New York, NY, USA},
doi = {10.1145/1143844.1143859},
booktitle = {Proceedings of the 23rd International Conference on Machine Learning},
pages = {113–120},
numpages = {8},
location = {Pittsburgh, Pennsylvania, USA},
series = {ICML '06}
}

@inproceedings{NIPS2009_0d0871f0,
 author = {Wallach, Hanna and Mimno, David and McCallum, Andrew},
 booktitle = {Advances in Neural Information Processing Systems},
 editor = {Y. Bengio and D. Schuurmans and J. Lafferty and C. Williams and A. Culotta},
 publisher = {Curran Associates, Inc.},
 title = {Rethinking LDA: Why Priors Matter},
  volume = {22},
 year = {2009}
}

@inproceedings{10.5555/1699510.1699543,
author = {Ramage, Daniel and Hall, David and Nallapati, Ramesh and Manning, Christopher D.},
title = {Labeled LDA: a supervised topic model for credit attribution in multi-labeled corpora},
year = {2009},
isbn = {9781932432596},
publisher = {Association for Computational Linguistics},
address = {USA},
booktitle = {Proceedings of the 2009 Conference on Empirical Methods in Natural Language Processing: Volume 1 - Volume 1},
pages = {248–256},
numpages = {9},
location = {Singapore},
series = {EMNLP '09}
}

@article{10.1145/3462478,
author = {Chauhan, Uttam and Shah, Apurva},
title = {Topic Modeling Using Latent Dirichlet allocation: A Survey},
year = {2021},
issue_date = {September 2022},
publisher = {Association for Computing Machinery},
address = {New York, NY, USA},
volume = {54},
number = {7},
issn = {0360-0300},
doi = {10.1145/3462478},
journal = {ACM Comput. Surv.},
month = {sep},
articleno = {145},
numpages = {35}
}

@article{Jelodar2017LatentDA,
  title={Latent Dirichlet allocation (LDA) and topic modeling: models, applications, a survey},
  author={Hamed Jelodar and Yongli Wang and Chi Yuan and Xia Feng},
  journal={Multimedia Tools and Applications},
  year={2017},
  volume={78},
  pages={15169 - 15211},
 }

@article{Gibbs,
author = {Geman, Stuart and Geman, Donald},
year = {1984},
month = {11},
pages = {721-741},
title = {Geman, D.: Stochastic relaxation, Gibbs distribution, and the Bayesian restoration of images. IEEE Trans. Pattern Anal. Mach. Intell. PAMI-6(6), 721-741},
volume = {6},
journal = {IEEE Trans. Pattern Anal. Mach. Intell.},
doi = {10.1109/TPAMI.1984.4767596}
}

@inproceedings{NIPS2013_9aa42b31,
 author = {Mikolov, Tomas and Sutskever, Ilya and Chen, Kai and Corrado, Greg S and Dean, Jeff},
 booktitle = {Advances in Neural Information Processing Systems},
 editor = {C.J. Burges and L. Bottou and M. Welling and Z. Ghahramani and K.Q. Weinberger},
 publisher = {Curran Associates, Inc.},
 title = {Distributed Representations of Words and Phrases and their Compositionality},
  volume = {26},
 year = {2013}
}

@InProceedings{pmlr-v48-miao16,
  title = 	 {Neural Variational Inference for Text Processing},
  author = 	 {Miao, Yishu and Yu, Lei and Blunsom, Phil},
  booktitle = 	 {Proceedings of The 33rd International Conference on Machine Learning},
  pages = 	 {1727--1736},
  year = 	 {2016},
  editor = 	 {Balcan, Maria Florina and Weinberger, Kilian Q.},
  volume = 	 {48},
  series = 	 {Proceedings of Machine Learning Research},
  address = 	 {New York, New York, USA},
  month = 	 {20--22 Jun},
  publisher =    {PMLR},
}

@article{Kingma2013AutoEncodingVB,
  title={Auto-Encoding Variational Bayes},
  author={Diederik P. Kingma and Max Welling},
  journal={CoRR},
  year={2013},
  volume={abs/1312.6114},
 }

@inproceedings{Cao2015ANN,
  title={A Novel Neural Topic Model and Its Supervised Extension},
  author={Ziqiang Cao and Sujian Li and Yang Liu and Wenjie Li and Heng Ji},
  booktitle={AAAI Conference on Artificial Intelligence},
  year={2015},
 }

@article{Moody2016MixingDT,
  title={Mixing Dirichlet Topic Models and Word Embeddings to Make lda2vec},
  author={Christopher E. Moody},
  journal={ArXiv},
  year={2016},
  volume={abs/1605.02019},
  }

@inproceedings{Srivastava2017AutoencodingVI,
  title={Autoencoding Variational Inference For Topic Models},
  author={Akash Srivastava and Charles Sutton},
  booktitle={International Conference on Learning Representations},
  year={2017},
  }

@article{Angelov2020Top2VecDR,
  title={Top2Vec: Distributed Representations of Topics},
  author={Dimitar Angelov},
  journal={ArXiv},
  year={2020},
  volume={abs/2008.09470},
 }

@article{7676413,
  author={Zhao, Renbo and Tan, Vincent Y. F.},
  journal={IEEE Transactions on Signal Processing}, 
  title={Online Nonnegative Matrix Factorization With Outliers}, 
  year={2017},
  volume={65},
  number={3},
  pages={555-570},
  doi={10.1109/TSP.2016.2620967}}

@inproceedings{murugaraj-etal-2025-mining,
    title = "Mining the Past: A Comparative Study of Classical and Neural Topic Models on Historical Newspaper Archives",
    author = "Murugaraj, Keerthana  and
      Lamsiyah, Salima  and
      During, Marten  and
      Theobald, Martin",
    editor = {H{\"a}m{\"a}l{\"a}inen, Mika  and
      {\"O}hman, Emily  and
      Bizzoni, Yuri  and
      Miyagawa, So  and
      Alnajjar, Khalid},
    booktitle = "Proceedings of the 5th International Conference on Natural Language Processing for Digital Humanities",
    month = may,
    year = "2025",
    address = "Albuquerque, USA",
    publisher = "Association for Computational Linguistics",
    pages = "452--463",
    ISBN = "979-8-89176-234-3"
}

@misc{grootendorst2020keybert,
  author       = {Maarten Grootendorst},
  title        = {KeyBERT: Minimal keyword extraction with BERT.},
  year         = 2020,
  publisher    = {Zenodo},
  version      = {v0.3.0},
  doi          = {10.5281/zenodo.4461265},
 }

@inproceedings{10.5555/3042817.3042832,
author = {Bergstra, J. and Yamins, D. and Cox, D. D.},
title = {Making a science of model search: hyperparameter optimization in hundreds of dimensions for vision architectures},
year = {2013},
publisher = {JMLR.org},
booktitle = {Proceedings of the 30th International Conference on International Conference on Machine Learning - Volume 28},
pages = {I–115–I–123},
location = {Atlanta, GA, USA},
series = {ICML'13}
}

@inproceedings{yang-etal-2011-topic,
    title = "Topic Modeling on Historical Newspapers",
    author = "Yang, Tze-I  and
      Torget, Andrew  and
      Mihalcea, Rada",
    editor = "Zervanou, Kalliopi  and
      Lendvai, Piroska",
    booktitle = "Proceedings of the 5th {ACL}-{HLT} Workshop on Language Technology for Cultural Heritage, Social Sciences, and Humanities",
    month = jun,
    year = "2011",
    address = "Portland, OR, USA",
    publisher = "Association for Computational Linguistics",
    url = "https://aclanthology.org/W11-1513/",
    pages = "96--104"
}

@inproceedings{6a042974c6ec4a4494c81711b929348b,
  title        = {Historizing Topic Models: A Distant Reading of Topic Modeling Texts within Historical Studies},
  author       = {Mats Fridlund and René Brauer},
  booktitle    = {Cultural Research in the Context of Digital Humanities},
  editor       = {L.V. Nikiforova and N.V. Nikiforova},
  year         = {2013},
  pages        = {152--163},
  publisher    = {Herzen State Pedagogical University},
  address      = {Russian Federation},
  language     = {English}
}

@inproceedings{medvecki2023multilingual,
  title={Multilingual transformer and bertopic for short text topic modeling: The case of serbian},
  author={Medvecki, Darija and Ba{\v{s}}aragin, Bojana and Ljaji{\'c}, Adela and Milo{\v{s}}evi{\'c}, Nikola},
  booktitle={Conference on Information Technology and its Applications},
  pages={161--173},
  year={2023},
  organization={Springer}
}

@inproceedings{mutsaddi-etal-2025-bertopic,
    title = "{BERT}opic for Topic Modeling of {H}indi Short Texts: A Comparative Study",
    author = "Mutsaddi, Atharva  and
      Jamkhande, Anvi  and
      Thakre, Aryan Shirish  and
      Haribhakta, Yashodhara",
    editor = "Weerasinghe, Ruvan  and
      Anuradha, Isuri  and
      Sumanathilaka, Deshan",
    booktitle = "Proceedings of the First Workshop on Natural Language Processing for Indo-Aryan and Dravidian Languages",
    month = jan,
    year = "2025",
    address = "Abu Dhabi",
    publisher = "Association for Computational Linguistics",
    url = "https://aclanthology.org/2025.indonlp-1.3/",
    pages = "22--32",
   
}

@article{ArsenievKoehler2020IntegratingTM,
  title={Integrating topic modeling and word embedding to characterize violent deaths},
  author={Alina Arseniev-Koehler and Susan D. Cochran and Vickie M. Mays and Kai Wei Chang and Jacob Gates Foster},
  journal={Proceedings of the National Academy of Sciences of the United States of America},
  year={2020},
  volume={119},
  url={https://api.semanticscholar.org/CorpusID:235658060}
}

@inproceedings{Cvejoski2023NeuralDF,
  title={Neural Dynamic Focused Topic Model},
  author={Kostadin Cvejoski and Rams{\'e}s J. S{\'a}nchez and C. Ojeda},
  booktitle={AAAI Conference on Artificial Intelligence},
  year={2023},
  url={https://api.semanticscholar.org/CorpusID:252564193}
}

@inproceedings{martinelli-etal-2024-exploring,
    title = "Exploring Neural Topic Modeling on a Classical {L}atin Corpus",
    author = "Martinelli, Ginevra  and
      Impiccich{\'e}, Paola  and
      Fersini, Elisabetta  and
      Mambrini, Francesco  and
      Passarotti, Marco",
    editor = "Calzolari, Nicoletta  and
      Kan, Min-Yen  and
      Hoste, Veronique  and
      Lenci, Alessandro  and
      Sakti, Sakriani  and
      Xue, Nianwen",
    booktitle = "Proceedings of the 2024 Joint International Conference on Computational Linguistics, Language Resources and Evaluation (LREC-COLING 2024)",
    month = may,
    year = "2024",
    address = "Torino, Italia",
    publisher = "ELRA and ICCL",
    url = "https://aclanthology.org/2024.lrec-main.606",
    pages = "6929--6934",

}

@article{ginn2024historia,
  title={Historia Magistra Vitae: Dynamic Topic Modeling of Roman Literature using Neural Embeddings},
  author={Ginn, Michael and Hulden, Mans},
  journal={arXiv preprint arXiv:2406.18907},
  year={2024}
}

 \end{document}